\documentclass[letterpaper]{article} 
\usepackage{aaai24}  
\usepackage{times}  
\usepackage{helvet}  
\usepackage{courier}  
\usepackage[hyphens]{url}  
\usepackage{graphicx} 
\usepackage{natbib}  
\usepackage{caption} 
\usepackage{algorithm}
\usepackage{algorithmic}

\usepackage{newfloat}
\usepackage{listings}
\DeclareCaptionStyle{ruled}{labelfont=normalfont,labelsep=colon,strut=off} 
\floatstyle{ruled}
\newfloat{listing}{tb}{lst}{}
\floatname{listing}{Listing}
\usepackage{amssymb}
\usepackage{tabularx}
\usepackage{multirow}
\newcolumntype{Y}{>{\centering\arraybackslash}X}
\usepackage{pifont}
\usepackage{amsmath}
\usepackage[svgnames]{xcolor}
\newcommand{\greencheck}{{\checkmark}}
\newcommand{\xmark}{\ding{55}}
\newcommand{\redcross}{{\xmark}}

\title{DeblurSR: Event-Based Motion Deblurring Under the Spiking Representation}
\author{
    Chen Song, Chandrajit Bajaj, Qixing Huang
}
\affiliations{
    The University of Texas at Austin\\

    Austin, TX 78712, USA\\
    \{song, bajaj, huangqx\}@cs.utexas.edu
}

\begin{document}

\maketitle

\begin{abstract}
We present DeblurSR, a novel motion deblurring approach that converts a blurry image into a sharp video. DeblurSR utilizes event data to compensate for motion ambiguities and exploits the spiking representation to parameterize the sharp output video as a mapping from time to intensity. Our key contribution, the Spiking Representation (SR), is inspired by the neuromorphic principles determining how biological neurons communicate with each other in living organisms. We discuss why the spikes can represent sharp edges and how the spiking parameters are interpreted from the neuromorphic perspective. DeblurSR has higher output quality and requires fewer computing resources than state-of-the-art event-based motion deblurring methods. We additionally show that our approach easily extends to video super-resolution when combined with recent advances in implicit neural representation.
\end{abstract}

\section{Introduction}
\label{Sec:Introduction}
\begin{figure}
    \centering
    \includegraphics[width=\columnwidth]{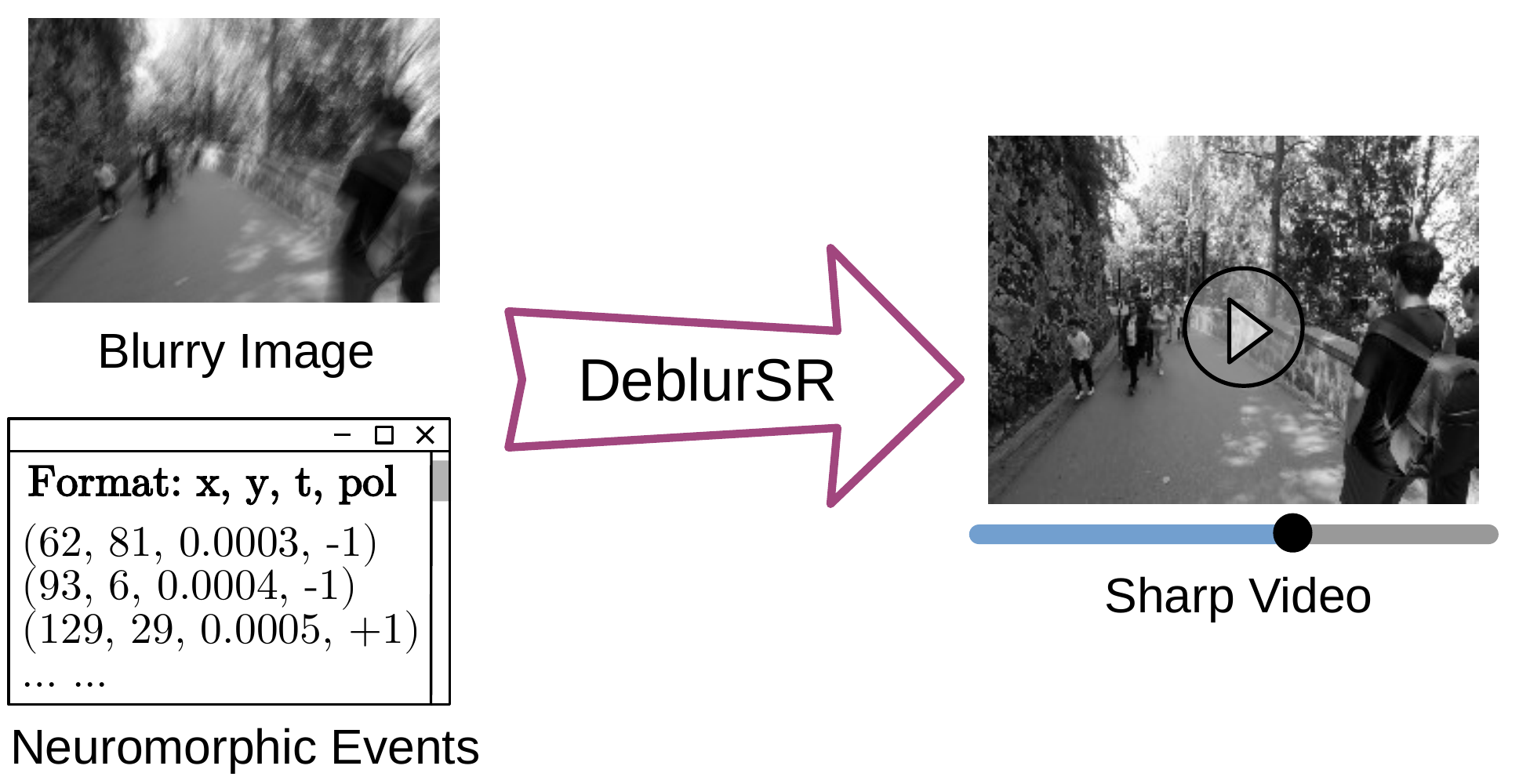}
    \caption{Problem Description. Imagine we are holding a camera while walking on the street. Relative movement between the camera and its surroundings causes motion blur, a visual artifact commonly observed by unprofessional photographers. With the help of neuromorphic events, which are a list of 4D points describing the coordinates, time, and polarity of intensity changes, DeblurSR converts the blurry image we take into a sharp video describing the camera's motion trajectory during the exposure interval.}
    \label{Fig:Problem}
\end{figure}

Neuromorphic events are commonly used in deblurring algorithms~\cite{pan2019bringing, pan2020high, jiang2020learning, lin2020learning, wang2020event, shang2021bringing, zhang2021fine, han2021evintsr, xu2021motion, sun2022event, kim2021event, song2022cir, wang2019event}. 
Modern neuromorphic devices are extremely fast and capture up to one million events per second. The enormous density of neuromorphic events has been shown to enable motion-deblurring algorithms to reverse the exposure process and recover the relative movement between the camera and the environment from one single image~\cite{pan2019bringing, wang2020event, song2022cir}. Figure~\ref{Fig:Problem} presents an illustration of the event-based image-to-video motion deblurring task.

While early motion-deblurring works apply numerical optimization techniques to directly solve for the sharp output video~\cite{pan2019bringing, pan2020high, wang2019event}, recent approaches utilize different data-driven pipelines as the inference model~\cite{wang2020event, song2022cir}. Despite these attempts, it is yet unclear how to properly instill prior knowledge about the neuromorphic event mechanism to build an effective deep learning paradigm that simultaneously addresses the motion ambiguity and emphasizes sharp edges. 

One interesting yet under-explored solution is to approximate the sharp output video by per-pixel parametric mappings from time to intensity and use deep learning to regress the parametric coefficients~\cite{song2022cir}. This allows the algorithm to fully exploit the speed of event cameras because the output theoretically has an infinitely high frame rate. However, common parametric kernels such as polynomial and trigonometric functions inherently assume that the underlying intensity signal is smooth and continuous. In reality, a sharp video contains numerous visual features that strongly contrast the background. The movement of a white object before a black background leads to instantaneous intensity flips between the two most extreme pixel values. It creates discontinuities in the intensity signal, which polynomials and trigonometric functions cannot represent. As the only existing work that uses per-pixel parametric mappings to represent videos with arbitrarily high frame rates, E-CIR~\cite{song2022cir} relies on a refinement module independent of the continuous parameterization to compensate for its limited representation capacity, resulting in a monolithic inefficient two-stage pipeline. 

\begin{figure}
    \centering
    \includegraphics[width=\columnwidth]{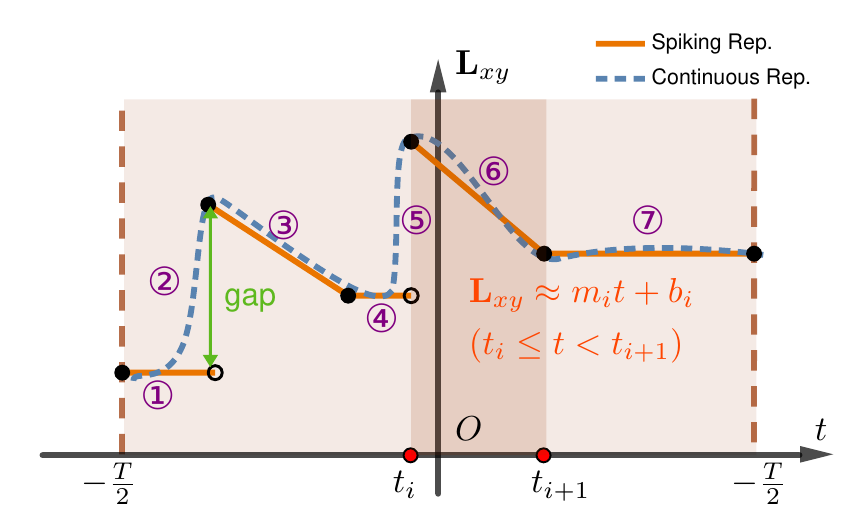}
    \caption{The Spiking Representation. Let $\mathbf{L}_{xy}(t)$ be the intensity of pixel $(x, y)$ at time $t$. $\mathbf{L}_{xy}(t)$ is approximated by a piece-wise linear function, analogous to the membrane potential of the neuron on the retina dedicated to pixel $(x, y)$. \ding{172} \ding{175} \ding{178}: The membrane potential does not change when the neuron is at rest. The pixel's intensity stays constant in the absence of motion. \ding{173} \ding{176}: Spikes occur at segment endpoints. The gap resembles the strength of the spike, which captures edge features caused by instantaneous changes. \ding{174} \ding{177}: The slope simulates the rate of membrane potential leakage in a biological neuron after its excitement, enabling the spiking representation to model gradual intensity changes. While the spiking representation (orange) captures signal discontinuities, the continuous parameterization (blue) does not.}
    \label{fig:spikes}
\end{figure}

As shown in Figure~\ref{fig:spikes}, we use a novel \textit{Spiking Representation (SR)} to approximate the per-pixel mappings from time to intensity. Imagine there is a dedicated visual receptor neuron for each pixel on the human retina. We model the light intensity of each pixel as the membrane potential of the corresponding receptor neuron, defined as the voltage difference between the interior and the exterior of the neuron cell. The spiking representation exploits a piece-wise linear function similar to the one developed by neuroscientists in the 1900s under the name of the Leaky-Integrate-and-Fire model~\cite{lapicque1907recherches}. Each linear segment's slope characterizes the gradual intensity change rate over time. The gaps at segment endpoints represent abrupt intensity changes caused by the edge movements. The spiking parameters, including the slope and intercept of each line segment and the segment endpoints, are predicted by a deep neural network. We further integrate the per-pixel piece-wise linear parameterization with spatial convolutional kernels, encouraging information to propagate among neighboring pixels. We refer to the proposed intensity modeling as the spiking representation because it mimics the spiking mechanism in biological neurons. 

We conduct extensive experimental analysis on the REDS dataset~\cite{Nah_2019_CVPR_Workshops_REDS} with synthetic events and the HQF dataset~\cite{stoffregen2020reducing} with real events. DeblurSR improves state-of-the-art reconstruction quality by 12.3\% on REDS and 22.2\% on HQF, while requiring a shorter training time and fewer computing resources. Through ablation studies, we demonstrate the strengths of three optional components, including the automatic keypoint selection module, the integral normalization constant, and the kernel-based spatial modeling. Finally, we show how DeblurSR naturally extends to video super resolution, improving the state-of-the-art method by 7.8\%.

In summary, we make the following contributions:
\begin{itemize}
\item We propose a novel Spiking Representation (SR) to parameterize sharp videos. We discuss the similarity between the spikes and the biological neural mechanism. We explain how the gaps in the spiking representation can be used to capture sharp edges. 

\item We train a deep network that regresses the spiking parameters from a blurry image and its associated events during the exposure interval. We show how to use the predicted parameters to render sharp videos.

\item In addition to motion deblurring, we extend the spiking representation to support video super resolution.

\end{itemize}

\section{Related Work}
\label{Sec:RelatedWork}

\subsection{Event Cameras and Spiking Neural Networks}
\label{Sec:RelatedWork:EventCameras}
Event cameras~\cite{4444573, mahowald1992vlsi} react to intensity changes and produce a stream of \textit{events}. Each event is represented as a 4-tuple $(x, y, t, p)$, including the pixel coordinates, the timestamp, and the binary polarity of the intensity change. Modern event cameras are able to detect up to one million events per second, allowing them to capture extremely fast motion details. However, events are noisy and unable to account for the exact magnitudes of the intensity changes, leading to challenges in algorithm design. A thorough introduction to the technical properties of event cameras is beyond the scope of this paper. We refer interested readers to Gallego et al.~(\citeyear{9138762}) for a comprehensive survey.

Spiking Neural Networks (SNNs)~\cite{MAASS19971659} are popular tools to build an event-based vision system because both event cameras and SNNs simulate how biological neurons behave. Different mathematical models are constructed to explain the spiking mechanism~\cite{lapicque1907recherches, dutta2017leaky, hodgkin1952quantitative, izhikevich2003simple, borisyuk1997information}. Our work is particularly inspired by the Leaky-Integrate-and-Fire (LIF) model~\cite{lapicque1907recherches}, where each neuron has a membrane potential whose value decays linearly with time in the logarithmic space and plunges abruptly whenever there is a spike. While prior works in event vision exploit SNNs as a black-box inference network~\cite{o2013real, 7280696, esser2016convolutional, rueckauer2017conversion, zhang2021event, zhu2022event, zhang2022spiking, lee2022fusion}, we extend the LIF model to directly model intensity changes in a video under the spiking representation. For readers familiar with SNNs, the most significant difference between SNNs and DeblurSR is that SNNs implicitly store the membrane potential and explicitly output the spikes, whereas DeblurSR explicitly models the membrane potential of visual receptor neurons and implicitly represents the spikes through parameterized coefficients.

\subsection{Motion Deblurring}
\label{Sec:RelatedWork:MotionDeblurring}

Motion deblurring algorithms turn a blurry image into a sharp video, allowing humans and downstream vision applications to understand movements during the image formation process. Motion deblurring is an ill-posed problem because the blurry image alone fails to capture critical motion parameters such as the moving direction and speed~\cite{jin2018learning, purohit2019bringing}. In Figure~\ref{Fig:Problem}, it is impossible to tell whether the photographer is moving forward into the woods or backward out of the screen from the blurry image itself. Thanks to the fast data rate of event cameras, several prior works utilize event streams to supplement the blurry input image to address the motion ambiguity~\cite{pan2019bringing, pan2020high, jiang2020learning, lin2020learning, wang2020event, shang2021bringing, zhang2021fine, han2021evintsr, xu2021motion, tulyakov2021time, Paikin_2021_CVPR, tulyakov2022time, zhang2022unifying, song2022cir, sun2022event, kim2021event, wang2019event} in not only image-to-video deblurring but also image-to-image deblurring and frame interpolation. Closely related to our design, E-CIR~\cite{song2022cir} represents pixel intensities in the output video as polynomial functions characterized by the temporal derivatives as selected event timestamps. This paper argues that the continuous nature of the polynomial representation limits the ability of E-CIR to generate sharp videos. On the other hand, the proposed spiking representation resembles the same biological principle that is followed by event cameras determining how neurons communicate with each other in living organisms, allowing DeblurSR to represent sharp discontinuous edges and enjoy strong interpretability.

\section{Method}
\label{Sec:Method}


\subsection{Event Camera Principles}
\label{Sec:Method:Principles}
On a pixel grid with resolution $h \times w$, let $\mathbf{L}_{xy}(t)$ be the intensity of pixel $(x, y)$ at time $t$. In the natural logarithmic space, consider the amount of intensity change from the previous timestamp $t'$ to the current timestamp $t$:
\begin{equation}
    \Delta ln[\mathbf{L}_{xy}(t)] = ln[\mathbf{L}_{xy}(t)] - ln[\mathbf{L}_{xy}(t')]
\end{equation}

At time $t$, an \textit{event}, $(x, y, t, p)$, indicates that for pixel $(x, y)$, the instantaneous intensity change exceeds the event generation threshold:
\begin{align}
    \Delta ln[\mathbf{L}_{xy}(t)] \geq c^+ & \text{, if } p = 1 \nonumber \\
    \Delta ln[\mathbf{L}_{xy}(t)] \leq c^- & \text{, if } p = -1 
    \label{eq:polarity-definition}
\end{align}
where $p = \pm1$ is the polarity of the event; $c^+$ and $c^-$ are event generation thresholds corresponding to positive events (intensity increments) and negative events (intensity decrements), respectively.

Modern event cameras simultaneously capture a stream of high-speed events and another stream of low-speed conventional frames. Let $\mathbf{B}$ be the $h \times w$ conventional frame captured during an exposure interval $[-\frac{T}{2}, \frac{T}{2}]$ with length $T$. Mathematically, the conventional frame is the temporal average of the true physical intensities:
\begin{equation}
    \mathbf{B}_{xy} = \frac{1}{T} \int_{-\frac{T}{2}}^{\frac{T}{2}} \mathbf{L}_{xy}(t) dt
    \label{eq:blurry-definition}
\end{equation}
During the exposure interval, relative motion between the camera and environment leads to undesirable blurriness in the conventional frame capture $\mathbf{B}$, which is a visual artifact commonly observed by unprofessional photographers.

\subsection{Problem Description}
\label{Sec:Method:Problem}
As illustrated by Figure~\ref{Fig:Problem}, the input to the event-based motion deblurring problem includes a blurry image $\mathbf{B} = \{\mathbf{B}_{xy}\}$ and all the events $\mathbf{E} = \{(x, y, t, p) | -\frac{T}{2} \leq t \leq \frac{T}{2}\}$ occurred during the exposure interval. The output of the problem is a sharp video with a temporal range of $[-\frac{T}{2}, \frac{T}{2}]$.

\subsection{Prediction Algorithm}
\label{Sec:Method:Algorithm}
\noindent\textbf{The Spiking Representation.} For each pixel $(x, y)$, we propose to approximate its latent intensity $\mathbf{L}_{xy}(t)$ as a parametric mapping from the temporal space $[-\frac{T}{2}, \frac{T}{2}]$ to the normalized intensity space $[0, 1]$. This contrasts with the conventional video representation, where $\mathbf{L}_{xy}(t)$ is characterized by discrete samples uniformly distributed across the exposure interval, and enjoys the advantage of having an infinitely high frame rate. Given a blurry image and its associated events, our algorithm learns to predict the coefficients of latent intensity parametric mappings for all the pixels. As shown in Figure~\ref{fig:spikes}, the proposed spiking representation uses disconnected line segments to approximate the per-pixel intensity mappings. Within each segment, the coefficients include the slope and the intercept. 

\begin{figure*}[t]
\centering
\includegraphics[width=0.95\textwidth]{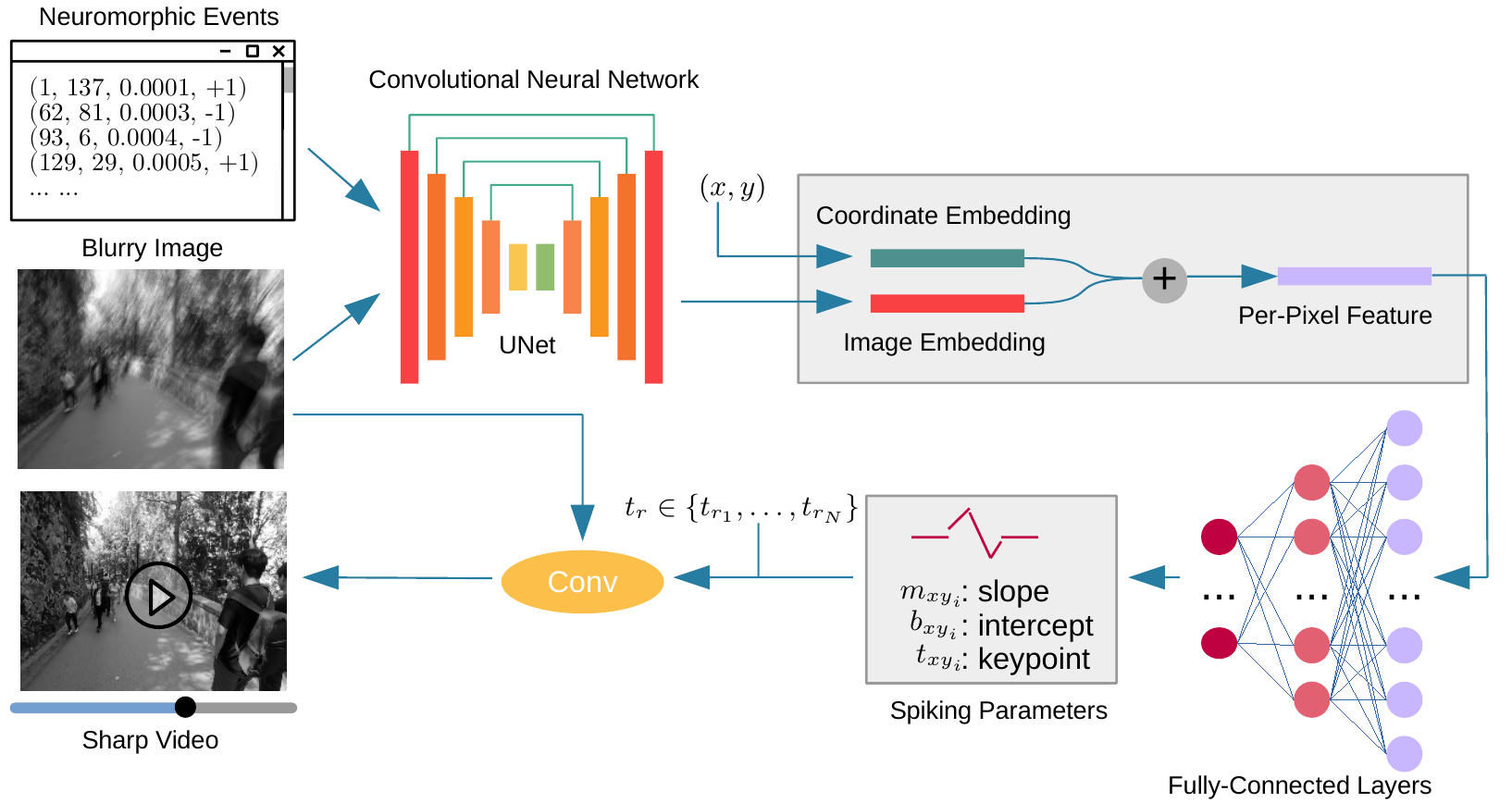}
\caption{The Overall Pipeline. Given a blurry image and its associated events in the exposure interval, we apply a Convolutional Neural Network (CNN) to extract an image embedding with the same spatial resolution as the input. For each pixel $(x, y)$, we fuse the image embedding with the coordinate embedding using the addition operation. A group of fully-connected layers take the resulting per-pixel feature vector as input and regress the spiking parameters for each pixel as output. At time $t_r$, we assemble a spatially varying kernel from the predicted spiking parameters. The convolution of this kernel with the input blurry image gives the output sharp frame at time $t_r$. By changing the timestamps, the spiking representation allows DeblurSR to render a sharp video with an arbitrarily high frame rate.}
\label{fig:model}
\end{figure*}

\noindent\textbf{Automatic Keypoint Selection Scheme.} As illustrated in Figure~\ref{fig:spikes}, a piecewise linear function with $n$ pieces has $n+1$ endpoints. In this paper, we use the term \textit{keypoints} to refer to the timestamps of these endpoints when the spikes happen. For each pixel $(x, y)$, the sharp intensity $\mathbf{L}_{xy}(t)$ is parameterized by $n$ slopes, $n$ intercepts, and $n+1$ keypoints:
\begin{equation}
    \mathbf{L}_{xy}(t) \approx \begin{cases}
    m_{{xy}_1}t + b_{{xy}_1} & t_{{xy}_1} \leq t < t_{{xy}_2} \\
    m_{{xy}_2}t + b_{{xy}_2} & t_{{xy}_2} \leq t < t_{{xy}_3} \\
    \dots \\
    m_{{xy}_n}t + b_{{xy}_n} & t_{{xy}_n} \leq t \leq t_{{xy}_{n+1}} \\
    \end{cases}
    \label{eq:spking-definition-1}
\end{equation}
Semantically, a keypoint represents a critical moment when the pixel's intensity changes significantly, which has a natural correlation to the event generation model described in Equation~(\ref{eq:polarity-definition}). However, the raw event data is highly irregular. During the exposure interval, different pixels have vastly different numbers of events, presenting a substantial challenge for efficient computation in parallel. Events are also known to be noisy~\cite{pan2019bringing, wang2020event, zhang2022unifying}. It is, therefore, inappropriate to directly extract the raw event timestamps as keypoints. To effectively exploit the correlation between events and keypoints while allowing efficient parallel computation, we propose the automatic keypoint selection module that is  robust against noise. Given the blurry image and its associated events, we employ a neural network to predict a set of $n$ segment widths for each pixel: $\{w_{{xy}_1}, w_{{xy}_2}, \dots, w_{{xy}_n}\}$. Through activation functions, we normalize the sum of these widths to the exposure length, $T$. The keypoints are $t_{{xy}_1} = -\frac{T}{2}$ and:
\begin{equation}
    t_{{xy}_i} = t_{{xy}_{i-1}} + w_{{xy}_{i-1}} \quad i = 2, 3, \dots, n+1
\end{equation}
An alternative to the automatic keypoint selection scheme is to use a heuristic-based algorithm to choose the keypoints from all event timestamps. This approach is similar to how E-CIR decides the Lagrange bases for the polynomial functions. 

\noindent\textbf{Integral Normalization Constant.} 
The basic version of the spiking representation discussed above may not satisfy the physical model in Equation~(\ref{eq:blurry-definition}) if the coefficients are predicted by a neural network. To fully utilize the information in the blurry input image, we amend the spiking representation in Equation~(\ref{eq:spking-definition-1}) with a normalization constant $c_{xy}$:
\begin{equation}
    \mathbf{L}_{xy}(t) \approx c_{xy} + \begin{cases}
    m_{{xy}_1}t + b_{{xy}_1} & t_{{xy}_1} \leq t < t_{{xy}_2} \\
    m_{{xy}_2}t + b_{{xy}_2} & t_{{xy}_2} \leq t < t_{{xy}_3} \\
    \dots \\
    m_{{xy}_n}t + b_{{xy}_n} & t_{{xy}_n} \leq t \leq t_{{xy}_{n+1}} \\
    \end{cases}
    \label{eq:spking-definition-2}
\end{equation}
where, given $m_{{xy}_i}$'s, $b_{{xy}_i}$'s, and $\mathbf{B}_{xy}$, the value of $c_{xy}$ can be solved analytically from Equation~(\ref{eq:blurry-definition}) by calculating the indefinite integral of Equation~(\ref{eq:spking-definition-2}):
\begin{align}
    c_{xy} =& -\mathbf{B}_{xy}
 + \frac{1}{2T} \sum_{i=1}^n m_{{xy}_i} (t_{{xy}_{i+1}}^2 - t_{{xy}_i}^2) + \\
& \frac{1}{T}\sum_{i=1}^n b_i (t_{{xy}_{i+1}} - t_{{xy}_i})
\end{align}

\noindent\textbf{Kernel-Based Spatial Modeling.} The sharpness of edges largely determines the quality of motion deblurring. While the above discussion addresses temporal intensity changes, it fails to incorporate spatial brightness variations, which is also key to edge sharpness. To address this issue, we predict $k^2$ different sets of spiking parameters for each pixel and use them to assemble a $k \times k$ convolutional kernel $\mathbf{K}_{xy}(t)$. After the spiking representation is further enhanced with spatial kernels, the sharp intensity is given as:
\begin{equation}
    \mathbf{L}_{xy}(t) \approx c_{xy} + \mathbf{K}_{xy}(t) \circledast \mathcal{N}(\mathbf{B}_{xy})
\end{equation}
where $\circledast$ stands for the inner product operator between two flattened vectors, and $\mathcal{N}(\mathbf{B}_{xy})$ is a $k \times k$ neighborhood for pixel $(x, y)$ in the input blurry image. 
We refer interested readers to our open-source GitHub repository for implementation details.

\noindent\textbf{Prediction Pipeline.} We present the overall motion deblurring pipeline in Figure~\ref{fig:model}. We first voxelize the irregular event input into an $m \times h \times w$ histogram tensor~\cite{Zhu_2019_CVPR}, where $m$ is the number of histogram bins. We then concatenate this histogram tensor with the blurry image, creating an input tensor whose dimensions are $(m+1) \times h \times w$. A UNet~\cite{ronneberger2015u}-based convolutional neural network takes this concatenated tensor as input and extracts a $d \times h \times w$ dimensional image embedding. Next, we utilize a linear layer to predict another $d$-dimensional coordinate embedding from the 2D coordinates of each pixel. We fuse the pixel-wise image and coordinate embeddings by the addition operation and employ fully-connected layers to regress the spiking parameters, including $k^2$ slopes, $k^2$ intercepts, and $n$ widths between $n+1$ keypoints for each pixel, where $k$ is the spatial kernel size. The fully connected layers have a total output size of $2k^2+n$ Finally, to render a sharp video with $N$ frames, we assemble the spatial kernels at rendering timestamps $ \{ t_{r_1}, \dots, t_{r_N} \}$ and convolve the kernels with the blurry image. The entire pipeline is trained end-to-end using the L1 loss on reconstructed sharp frames. 

\noindent\textbf{Extension to Super Resolution.} Thanks to coordinate embedding, DeblurSR can predict the spiking parameters for pixels with non-integer coordinates in between two regular pixels. During testing, the coordinates take non-integer values such as $(1.5, 7)$, representing the midpoint of two regular pixels. This allows natural support to video super resolution. Formally, given a blurry image and events defined on an $h \times w$ grid, the super-resolution problem aims at reconstructing a sharp video with a higher resolution $h' \times w'$.  In Section~\ref{Sec:Evaluation:Super}, we show that DeblurSR achieves state-of-the-art super-resolution performance even without any high-resolution training supervision.

\section{Evaluation}
\label{Sec:Evaluation}

\begin{figure*}
    \centering
    \includegraphics[width=\textwidth]{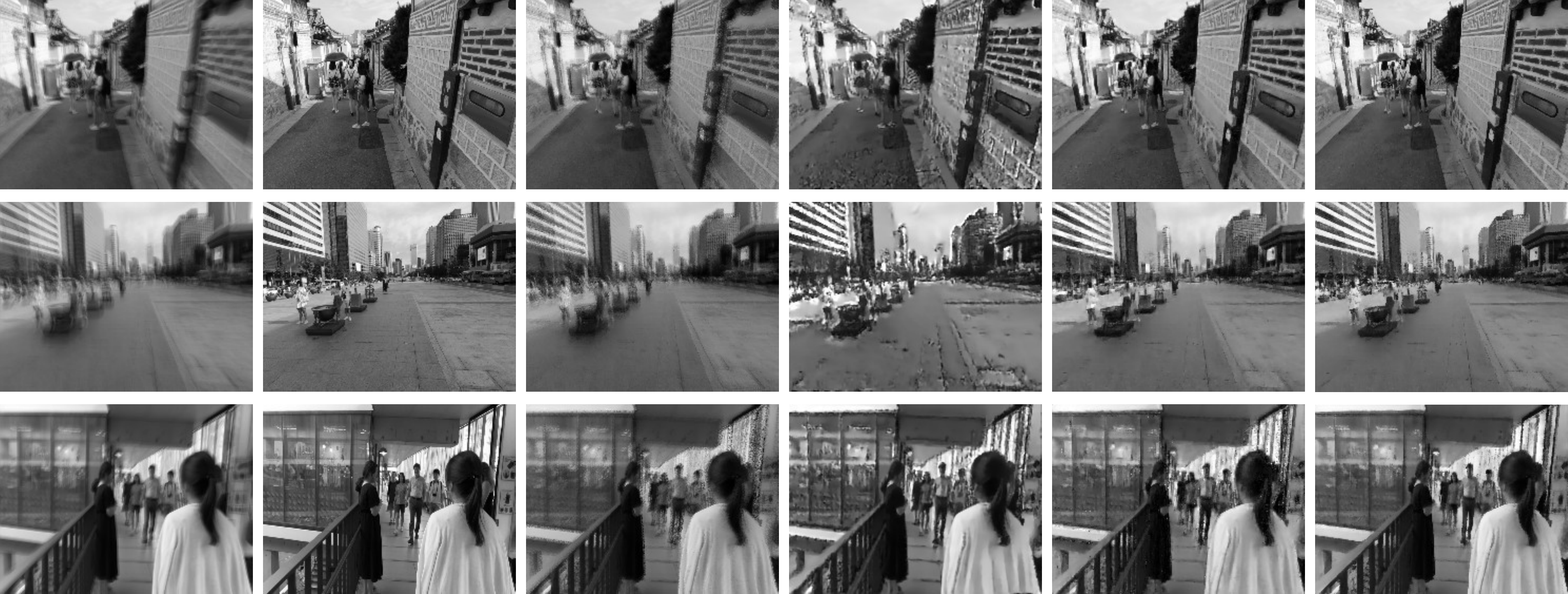}
    \begin{tabularx}{\textwidth}{Y Y Y Y Y Y}
        Input & Ground Truth & EDI & eSL-Net & E-CIR & Ours
    \end{tabularx}
    \caption{Visualizations on REDS. Video results are available on our GitHub page with higher illustration quality.}
    \label{fig:reds_1}
\end{figure*}

\begin{figure*}
    \centering
    \includegraphics[width=\textwidth]{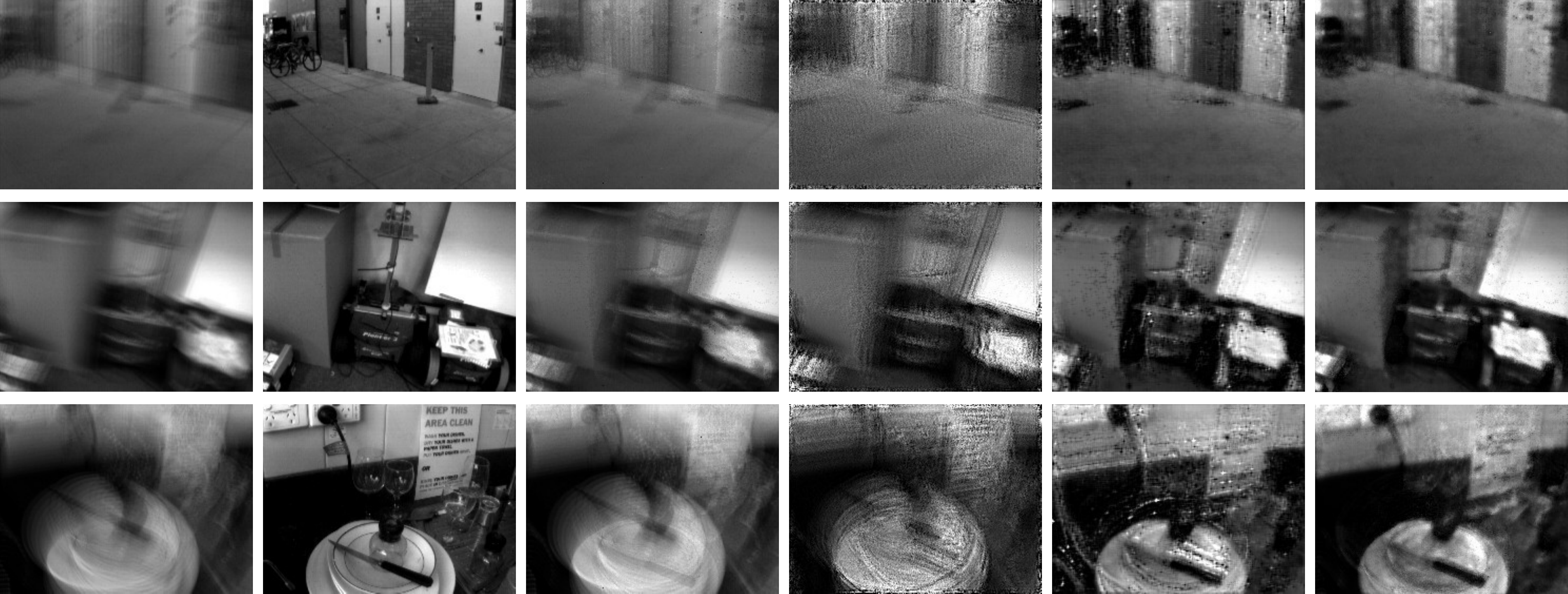}
    \begin{tabularx}{\textwidth}{Y Y Y Y Y Y}
        Input & Ground Truth & EDI & eSL-Net & E-CIR & Ours
    \end{tabularx}
    \caption{Visualizations on HQF. Video results are available on our GitHub page with higher illustration quality.}
    \label{fig:hqf}
\end{figure*}

\begin{table*}
    \begin{center}
        \begin{tabular}{c | c c c | c c c}
            \hline
            \multirow{2}{*}{Methods}  & \multicolumn{3}{c|}{Performance on REDS} & \multicolumn{3}{c}{Performance on HQF} \\
            \cline{2-7}
             & MSE $\downarrow$ & PSNR $\uparrow$ & SSIM $\uparrow$ & MSE $\downarrow$ & PSNR $\uparrow$ & SSIM $\uparrow$ \\
            \hline
            EDI & 0.182 & 21.663 & 0.664 & 0.336 & 17.822 & 0.515 \\
            eSL-Net & 0.203 & 20.640 & 0.601 & 0.452 & 14.938 & 0.282 \\
            E-CIR & 0.114 & 25.531 & 0.819 & 0.207 & 21.713 & 0.609 \\
            \hline
            Ours & \textbf{0.100 $\pm$ 0.001} & \textbf{26.725 $\pm$ 0.001} & \textbf{0.859 $\pm$ 0.001} & \textbf{0.161 $\pm$ 0.001} & \textbf{23.910 $\pm$ 0.008} & \textbf{0.693 $\pm$ 0.001} \\
            \hline
        \end{tabular}
    \end{center}
    \caption{On both the REDS and the HQF datasets, DeblurSR outperforms baseline approaches that represent videos by frames and polynomials in terms of the quantitative motion deblurring quality.}
    \label{tab:baseline}
\end{table*}

\begin{table}
    \begin{center}
        \begin{tabular}{c | c c c | c c}
            \hline
            \multirow{2}{*}{} & \multicolumn{3}{c|}{Modules} & \multicolumn{2}{c}{MSE $\downarrow$} \\
            \cline{2-6}
            & AK & IN & KS & REDS & HQF  \\
            \hline
            1 & \redcross & \redcross & \redcross  & 0.107 $\pm$ 0.001 & 0.236 $\pm$ 0.002 \\
            2 & \greencheck & \redcross & \redcross & 0.103 $\pm$ 0.001 & 0.229 $\pm$ 0.001 \\
            3 & \greencheck & \greencheck & \redcross & 0.102 $\pm$ 0.001 & 0.163 $\pm$ 0.001 \\
            \hline
            4 & \greencheck & \greencheck & \greencheck & \textbf{0.100 $\pm$ 0.001} & \textbf{0.161 $\pm$ 0.001} \\
            \hline
        \end{tabular}
    \end{center}
    \caption{We use ablation studies to demonstrate the strengths of the automatic keypoint selection scheme (AK), the integral normalization constant (IN), as well as kernel-based spatial modeling (KS). }
    \label{tab:reds_ablation}
\end{table}


\subsection{Experimental Setup}
\label{Section:Datasets}

We conduct experimental evaluations on two benchmark datasets. The REalistic and Dynamic Scenes (REDS)~\cite{Nah_2019_CVPR_Workshops_REDS} dataset is a popular dataset used to evaluate deblurring approaches. The original REDS dataset contains sharp videos with various real-world contents released under the CC BY 4.0 license. Following Song et al.~(\citeyear{song2022cir}), we use the ESIM simulator~\cite{Rebecq18corl} to synthesize events and blurry images. We then employ the official training and validation splits to train and test our model, respectively. Notably, this dataset is also referred to as the GroPro dataset by some authors~\cite{wang2020event}, although a much smaller dataset popular in image-to-image deblurring approaches happens to share the same name~\cite{Nah_2017_CVPR}. 

The High Quality Frames (HQF)~\cite{stoffregen2020reducing} dataset is another benchmark recently developed to evaluate event-based vision algorithms. The dataset is available for public download, but the licensing details are unclear. The HQF dataset contains both sharp videos and the associated event captures. We apply temporal averaging to generate blurry images from the sharp video. Following Zhang et al.~(\citeyear{zhang2022unifying}), we use five clips for testing and nine clips for training.

We compare DeblurSR to all the event-based image-to-video motion deblurring approaches with a complete (data preparation, training, and testing) open-source implementation known to us at the time of paper writing. We additionally evaluate eSL-Net~\cite{wang2020event} by creating customized training scripts for the incomplete code released on GitHub. 


\subsection{Training Details}
\label{Section:Details}

We implement DeblurSR under PyTorch~\cite{NEURIPS2019_9015} and utilize ADAM~\cite{kingma2014adam} to train the network for 50 epochs. We set the initial learning rate to 0.0001 and reduce the learning rate by half after 20 and 40 epochs, respectively. The number of line segments in the spiking representation is $n=10$. The dimension of spatial kernels is $k=3$. The number of histogram bins is $m=26$. The size of image and coordinate embeddings is $d=256$. More details of our experiments are available in the open-source GitHub repository.

\subsection{Motion Deblurring}
\label{Sec:Evaluation:Deblurring}
\noindent\textbf{Baseline Comparison.}
Table~\ref{tab:baseline} presents the quantitative evaluation on the REDS and HQF datasets. We use three image quality metrics to compare DeblurSR with different baseline approaches: the Mean Squared Error (MSE), the Peak Signal-to-Noise Ratio (PSNR), and the Structural Similarity Index Measure (SSIM). 

Our method demonstrates an impressive ability in motion deblurring. Specifically, on the REDS dataset, DeblurSR improves the current best-performing method by 12.3\% in MSE, 4.7\% in PSNR, and 4.9\% in SSIM. On HQF, DeblurSR outperforms the state-of-the-art approach by 22.2\% in MSE, 10.1\% in PSNR, and 14.0\% in SSIM.

Qualitatively, as shown in Figure~\ref{fig:reds_1} and Figure~\ref{fig:hqf}, DeblurSR generates smooth and sharp frames. In particular, we point out that our results are sharper than the EDI reconstruction, which assumes all events correspond to the same amount of intensity change. Meanwhile, our reconstructed frames are significantly less noisy than the eSL-Net reconstruction, which overly emphasizes texture details. Compared with E-CIR, DeblurSR offers more realistic details around the thin edges. The difference between E-CIR and DeblurSR is sometimes subtle and hard to notice from static images. 

\noindent\textbf{Efficiency.}
The proposed DeblurSR is computationally efficient. On REDS, it takes 100 hours to train E-CIR for 50 epochs using three Tesla V100 GPUs. By contrast, DeblurSR only requires 72 hours and two of the same GPUs under the identical training setting, representing a 28.0\% reduction in training time and a 33.3\% reduction in resource demand. The efficiency comes from the simplicity of our spiking representation in contrast with the high-order polynomial parameterization in E-CIR, which allows faster operations like computing the derivative and integral.

\noindent\textbf{Ablation Study.} Table~\ref{tab:reds_ablation} summarizes the quality of motion deblurring using different variants of the model. The comparison between the first and second rows shows that the proposed automatic keypoint selection module improves the heuristic-based keypoint selection algorithm in E-CIR by 3.7\% on REDS and 3.0\% on HQF. This result demonstrates the advantage of learning the critical timestamps from a large amount of training data.

From the second row to the third row, the integral normalization constant improves the deblurring quality by 1.0\% and 29.0\% on REDS and HQF, respectively. The improvement suggests that the physical model discussed in Equation~(\ref{eq:blurry-definition}) is an effective regularization for the neuromorphic spiking parameters. Noticeably, the improvement on HQF is a lot more significant than REDS. While REDS is a large synthetic dataset with 240 training clips, HQF is a small real benchmark with only 9 training clips. Empirically, we observe that the normalization constant is particularly beneficial when data is limited and the event noise is real and complex.

Finally, the last two rows in Table~\ref{tab:reds_ablation} examine the effectiveness of the spatial convolutional kernel modeling. The spatial kernel leads to 2.0\% and 0.6\% MSE improvement on REDS and HQF, respectively. Importantly, the kernel-based spatial modeling demonstrates that the spiking representation can also be used as an operator on the input and support complex structures. A promising future direction is to stack layers of such operators and construct a network.

\subsection{Super Resolution}
\label{Sec:Evaluation:Super}
\begin{table}[t]
    \begin{center}
        \begin{tabular}{r|c c c}
            \hline
            Methods & MSE $\downarrow$ & PSNR $\uparrow$ & SSIM $\uparrow$ \\
            \hline
            EDI (+bicubic) & 0.196 & 20.789 & 0.550 \\
            eSL-Net & 0.228 & 19.507 & 0.483 \\
            E-CIR (+bicubic) & 0.142 & 23.530 & 0.622 \\
            \hline
            Ours (LR supervision) & 0.140 & 23.642 & 0.634 \\
            Ours (HR supervision) & \textbf{0.131} & \textbf{24.272} & \textbf{0.664} \\
            \hline
        \end{tabular}
    \end{center}
    \caption{We extend DeblurSR to super-resolution and evaluate the performance on the REDS dataset.}
    \label{tab:reds_baseline_sr}
\end{table}

\begin{figure}[t]
    \centering
    \includegraphics[width=\columnwidth]{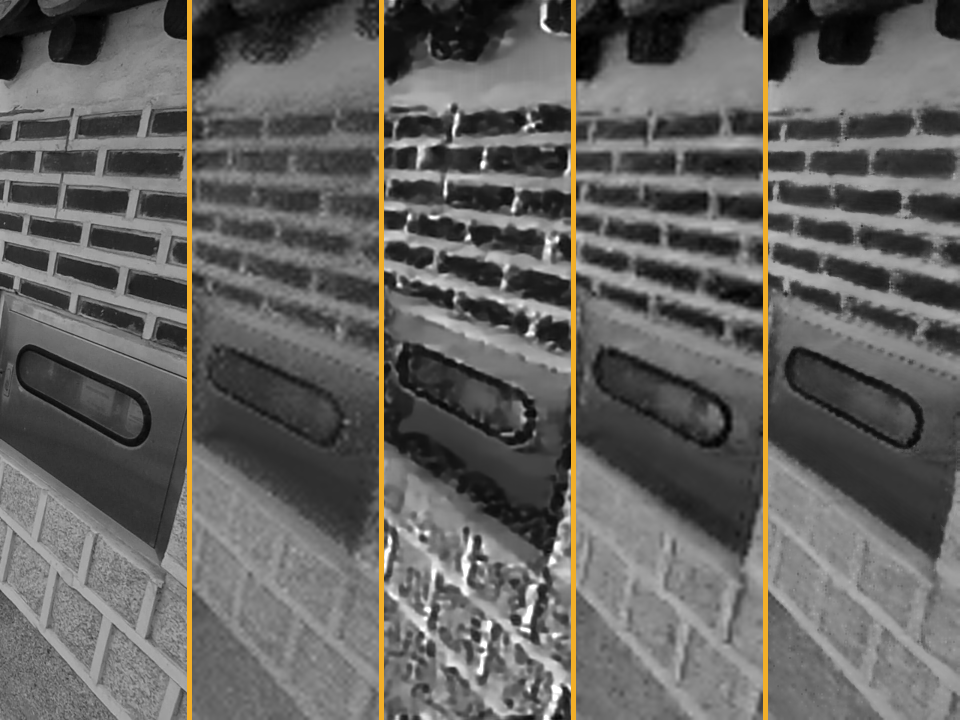}
    \begin{tabularx}{\columnwidth}{Y Y Y Y Y}
        (a) & (b) & (c) & (d) & (e)
    \end{tabularx}
    \caption{Visual Comparison Between Different Super-Resolution Methods. (a) Ground Truth; (b) EDI; (c) eSL-Net; (d) E-CIR; (e) Ours.} 
    \label{fig:sr_compare}
\end{figure}

As illustrated in Figure~\ref{fig:model}, DeblurSR naturally supports super resolution because the pixel coordinates $(x, y)$ can take non-integer values such as $(1.5, 7)$, representing the midpoint of two regular pixels. This section compares DeblurSR with three different baseline approaches on the REDS dataset. We provide the blurry image and the events in low resolution ($180\times240$) and evaluate the results in high resolution ($720\times960$). Among the three baseline approaches, we note that only eSL-Net uses ground-truth high-resolution frames in the training objective. Neither EDI nor E-CIR provides intrinsic support to super-resolution. We obtain high-resolution outputs from EDI and E-CIR through bicubic interpolation. From Table~\ref{tab:reds_baseline_sr}, we observe that DeblurSR achieves state-of-the-art performance with and without high-resolution supervision. In the absence of high-resolution supervision, DeblurSR improves E-CIR by 1.4\% in MSE. With high-resolution supervision, the relative improvement becomes 7.8\%. 
Figure~\ref{fig:sr_compare} further confirms our advantage through qualitative visualizations. 

\section{Conclusion}
\label{Sec:Conclusion}
In this paper, we introduce DeblurSR, a novel event-based motion deblurring approach based on the spiking representation. DeblurSR builds upon the same biological principles followed by the event camera design. Experiments show that DeblurSR outperforms state-of-the-art approaches in deblurring quality and can be easily extended to video super resolution. In the future, we plan to modify DeblurSR and allow different pixels to have a different number of parametric segments. This requires a non-trivial redesign of the prediction network to handle the heterogeneity. Another possible direction is to construct a general-purpose deep network with
layers of spiking neurons.

\section*{Acknowledgement}
\label{Sec:Acknowledgement}
This research was supported in part by a grant from the NIH DK129979, in part from the Peter O'Donnell Foundation, the Michael J Fox Foundation, Jim Holland-Backcountry Foundation, and in part by a grant from the Army Research Office accomplished under Cooperative Agreement Number W911NF-19-2-0333. Additionally, we acknowledge the
support from NSF Career IIS-2047677, NSF HDR-1934932, and NSF CCF-2019844.

\bibliography{aaai_final}

\end{document}